\documentclass[11pt]{article}

\usepackage[preprint]{acl}

\usepackage{times}
\usepackage{latexsym}
\usepackage[T1]{fontenc}
\usepackage[utf8]{inputenc}
\usepackage{microtype}
\usepackage{inconsolata}
\usepackage{graphicx}

\usepackage{amsmath}
\usepackage{amssymb}
\usepackage{booktabs}
\usepackage{multirow}
\usepackage{threeparttable}
\usepackage[table]{xcolor}
\usepackage{algorithm}
\usepackage{algorithmic}
\usepackage{newfloat}
\usepackage{listings}
\usepackage{tcolorbox}
\tcbuselibrary{breakable}
\DeclareCaptionStyle{ruled}{labelfont=normalfont,labelsep=colon,strut=off}
\floatstyle{ruled}
\newfloat{listing}{tb}{lst}{}
\floatname{listing}{Listing}
\newcommand{\projname}[0]{EviSeek}
\definecolor{draftblue}{RGB}{35,80,125}

\newcommand{\qwenname}{Qwen3-VL-32B}
\newcommand{\proname}{Gemini 3.1 Pro}

\definecolor{promptblue}{RGB}{47,91,146}
\definecolor{promptbackground}{RGB}{255,255,255}
\definecolor{prompttitle}{RGB}{224,236,251}
\newtcolorbox{promptbox}[1]{%
  colback=promptbackground,colframe=promptblue,
  colbacktitle=prompttitle,coltitle=black,
  breakable,width=\linewidth,lines before break=8,
  fonttitle=\normalsize\bfseries,halign title=flush left,title={#1},
  title after break={#1 (continued)},
  boxrule=0.9pt,arc=6pt,
  left=6pt,right=6pt,top=6pt,bottom=6pt,
  toptitle=4pt,bottomtitle=4pt,
  before skip=6pt,after skip=8pt}

\title{\projname{}: An Evidence-Centric Agentic Framework for Embodied Active Information Seeking}

\author{
  Jintao Li\textsuperscript{1,2,3},
  Zhiyong Cao\textsuperscript{1,2},
  Baihua Xiao\textsuperscript{1},
  Yu Bai\textsuperscript{3},
  B{\"o}rje F. Karlsson\textsuperscript{3}\\
  \textsuperscript{1}Institute of Automation, Chinese Academy of Sciences\\
  \textsuperscript{2}School of Artificial Intelligence, University of Chinese Academy of Sciences\\
  \textsuperscript{3}Beijing Academy of Artificial Intelligence
}

\begin{document}

\maketitle
\begin{abstract}
Active embodied question answering requires agents to acquire evidence through viewpoint changes and physical interaction.
The central challenge is to avoid undirected interaction: taking plausible actions without specifying what evidence they should reveal for answering the question.
Action plans and accumulated histories alone can leave the informational purpose of an interaction implicit.
We introduce \textbf{\textit{\projname{}}}, a training-free framework that makes question-derived \emph{evidence demands} explicit acquisition objectives, linking observation-conditioned actions and selectively retained evidence to the information the agent seeks.
On 3,436 ESI-Bench questions, EviSeek improves accuracy over the Single-model agent by 7.1 percentage points with \qwenname{} and 3.7 points with \proname{}, while improving visual-support-weighted accuracy and reducing mean token usage.
These findings support making the informational purpose of interaction explicit, with action selection and evidence retention organized around what the agent needs to learn to answer the question.
\end{abstract}

\section{Introduction}

Active embodied question answering (A-EQA) requires agents to acquire information through interaction with a partially observed environment~\citep{das2018embodied,gordon2017iqa0,majumdar2024openeqa}. Relevant evidence may be occluded, distributed across locations, or revealed only by viewpoint changes and physical intervention~\citep{bajcsy1988active,sawyer2024foundation}. Agents must decide what to observe, how to acquire those observations, and what to retain for answering.

ESI-Bench highlights \emph{action blindness}: poor action choices yield inadequate observations and subsequent reasoning errors~\citep{hong2026esibench}. We focus on three related difficulties: questions may leave the required observations unspecified; acquiring them may require several stages of interaction; and limited visual history may discard earlier evidence before answering. For example, assessing a stacking configuration requires observing its state after a physical test and retaining that outcome. A plausible manipulation sequence alone does not establish that the relevant evidence was acquired or preserved.

Existing agents guide interaction through histories, exploration, retrieval, and confidence~\citep{yao2022react,huang2022inner,ren2024explore,cheng2025efficienteqa}. ScoutVLA seeks missing visual evidence~\citep{lu2026scoutvla0}, MemoryEQA shares memory across planning, stopping, and answering~\citep{zhai2025memory}, and R4 decomposes questions for structured retrieval~\citep{sohn2026r4}. Building on these directions, we ask whether explicit \emph{evidence demands} can serve as a shared reference for acquisition and retention, improving active question answering with a fixed pretrained vision-language model (VLM).

We introduce \textbf{\textit{\projname{}}}, a training-free framework organized around this representation. \textbf{Evidence-Demand Planning} derives observational targets from the question and initial view without assuming a candidate answer is correct, then assigns acquisition subtasks to them. \textbf{Evidence-Seeking Interaction} pursues these goals through viewpoint changes or physical intervention, conditioning actions on observations and execution feedback. \textbf{Evidence-Oriented Memory} selectively retains views and action outcomes on an evidence board, linked to their originating subtask and demand. Subtasks organize acquisition, while the board separates retained answer evidence from transient control observations. Spatial reasoning and evidence interpretation remain dependent on the pretrained VLM.

On 3,436 ESI-Bench questions~\citep{hong2026esibench}, EviSeek improves overall accuracy over the official Single-model active agent by 7.1 percentage points with \qwenname{} and 3.7 with \proname{}, under the same task interface and environment-action ceiling. Both backbones also improve visual-support-weighted accuracy while using fewer model calls and tokens on average. Qwen ablations show that removing demands while retaining task decomposition lowers online accuracy. On fixed trajectories, replacing the board with recent-view memory or replacing selective retention with uniform sampling under the same frame-capacity limit also lowers accuracy. These comparisons support explicit demands and selective retention within the evaluated framework. A navigation variant improves grounded correctness over the RGB Single-model baseline on EXPRESS-Bench~\citep{jiang2025beyond}. Benefits remain task-dependent: Metric Comparison regresses with both ESI-Bench backbones, delimiting the gains from evidence organization.

Our main contributions are as follows:
\begin{itemize}
\item We introduce a demand-indexed representation linking question-derived observational targets, acquisition subtasks, and retained evidence.

\item We develop EviSeek, a training-free framework that uses this representation to organize viewpoint acquisition, physical intervention, and selective evidence retention.

\item We report gains in overall accuracy and visual-support-weighted accuracy with lower mean model usage on two ESI-Bench backbones, supported by demand and memory ablations, and evaluate a navigation variant on EXPRESS-Bench.
\end{itemize}

\section{Related Work}

\paragraph{Active Information Acquisition.}
Active preference inference~\citep{piriyakulkij2023active}, Active Task Disambiguation~\citep{kobalczyk2025active}, BED-LLM~\citep{choudhury2025bedllm}, and CA-BED~\citep{arnould2026bed} select informative queries to reduce uncertainty. Active perception studies viewpoint selection~\citep{zhang2024agent3d0zero0} and stopping through conformal prediction~\citep{ren2024explore} or image--question relevance~\citep{cheng2025efficienteqa}. ScoutVLA~\citep{lu2026scoutvla0} infers missing visual evidence and trains a dual-expert model for continuous UAV viewpoint refinement and terminal-view answering. EviSeek uses explicit demands to link acquisition subtasks with selectively retained observations under a frozen VLM, supporting both viewpoint changes and physical tests.

\paragraph{Spatial Navigation.}
Navigation agents use explicit reasoning~\citep{zhou2023navgpt0} and expert discussion~\citep{long2023discuss}, with RGB-video policies offering a map-free alternative~\citep{zhang2024navid}. Maps~\citep{huang2023visual,yokoyama2024vlfm,sakamoto2024map0based}, scene graphs~\citep{saxena2024grapheqa,ginting2025enter}, and visual memories~\citep{yang20243d0mem0} support exploration; Fine-EQA combines frontier search with goal-oriented inspection~\citep{jiang2025beyond}. MemoryEQA~\citep{zhai2025memory} updates structured memories through viewpoint comparison and retrieves module-specific information for frontier planning, stopping, and answering. Training-free R4~\citep{sohn2026r4} decomposes questions into semantic, spatial, and temporal retrieval keys over a 4D database constructed from RGB, point clouds, and SLAM. EviSeek uses demands to guide acquisition of new evidence and its selective retention for answering, without external mapping.

\paragraph{Interactive Question Answering and Execution.}
IQA~\citep{gordon2017iqa0} couples navigation with object interaction. Early EQA further studies modular control, point-cloud perception, and multi-target reasoning~\citep{das2018neural,wijmans2019embodied,yu2019multi0target}. Grounded planners~\citep{song2022llm0planner0,rana2023sayplan0,bai2026egoactor0}, affordance-based skill selection~\citep{ahn2022can}, executable navigation programs~\citep{ziliotto2025tango}, and visual programs~\citep{gupta2022visual,surís2023vipergpt0} connect language to execution. Agents refine execution through feedback, reflection, or tree search~\citep{sun2023adaplanner0,shinn2023reflexion0,zhou2023language}. On ESI-Bench, ParallelWorld~\citep{chen2026parallelworld} searches simulated branches, SHAPER~\citep{wang2026self} evolves skills and context, and Spatial-Interactor~\citep{yao2026spatial} learns from interaction trajectories. EviSeek links question-derived demands to executed actions and selectively retained observations using a frozen backbone.

\section{Preliminaries}
\label{sec:preliminary}
We now formalize the A-EQA setting introduced above, following OpenEQA~\cite{majumdar2024openeqa}. An A-EQA instance is specified by $$\mathcal{I}=(Q,\mathcal{E},A^\star),$$ where $Q$ is the question, $A^\star$ is the ground-truth answer, and $\mathcal{E}$ is an interactive environment instantiated either in simulation or in the physical world. In response to an agent action, $\mathcal{E}$ transitions to a new latent state and returns a new egocentric visual observation that may reveal the perceptible consequences of that action. Formally, an interaction step is described by $$(x_{t+1},o_{t+1})\sim\mathcal{E}(\cdot\mid x_t,a_t),$$ where $x_t$ denotes the latent environment state, which is not directly observable to the agent. The information available to the policy is represented by $$M_t=\mathcal{R}_t(H_t),\ H_t=(o_0,a_0,o_1,\ldots,a_{t-1},o_t),$$ where $H_t$ is the complete interaction history and $\mathcal{R}_t$ denotes the history-access mechanism employed by the agent. Depending on the interaction horizon and model architecture, $\mathcal{R}_t$ may preserve the complete history or produce a reduced representation through truncation, selection, compression, or structured aggregation~\cite{das2018embodied,zhai2025memory,saxena2024grapheqa,zhang2026fast,sohn2026r4}. At each decision step, the agent selects either an environmental action or a final answer according to $$a_t\sim\pi(\cdot\mid Q,M_t),\ a_t\in\mathcal{G}_{\mathcal{E}}\cup\{\textsc{Answer}(\hat{A})\},$$ where $\mathcal{G}_{\mathcal{E}}$ denotes the action space supported by $\mathcal{E}$, ranging from discrete action primitives to continuous low-level controls. Independently of this implementation granularity, any action $a_t\in\mathcal{G}_{\mathcal{E}}$ can be abstracted by how it changes the evidence available to the agent as either \textbf{1)} \emph{viewpoint acquisition}, which changes the observation geometry by moving the camera while leaving the non-agent world state unchanged; or \textbf{2)} \emph{physical intervention}, which modifies the environment state through action and acquires evidence from the observable consequences.  If $a_t\in\mathcal{G}_{\mathcal{E}}$, the action is executed in $\mathcal{E}$ and the returned observation extends the subsequent interaction history; if $a_t=\textsc{Answer}(\hat{A})$, the agent outputs $\hat{A}$ and terminates the interaction.

\begin{figure*}[t]
\centering
\includegraphics[width=0.99\textwidth]{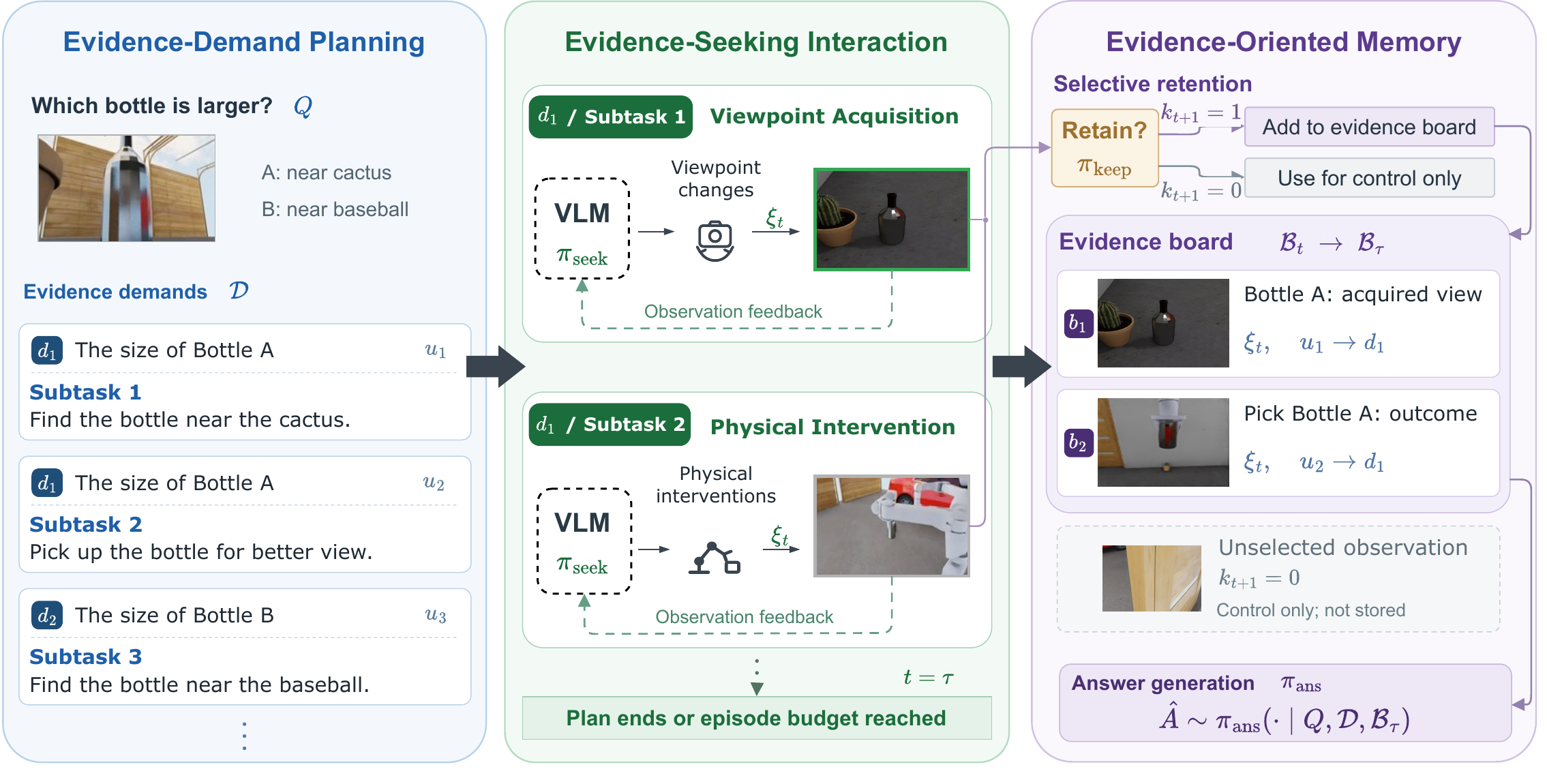}
\caption{EviSeek connects question-derived evidence demands to acquisition goals, observation-conditioned actions, and retained evidence for answering. Unselected observations support control but are excluded from the evidence board.}
\label{fig:method}
\end{figure*}

\section{Method}
\label{sec:method}
\subsection{Overview}

EviSeek formulates A-EQA as an evidence acquisition problem.
EviSeek represents the observations required by a question as evidence demands, then uses these demands to guide acquisition and retention.
The key idea is to maintain an evidence state that decouples \emph{what information is required}, \emph{how evidence is acquired}, and \emph{which observations are retained}.

At interaction step $t$, EviSeek instantiates the information state $M_t$ of Section~\ref{sec:preliminary} as a demand-indexed representation of the interaction history $H_t$:
\[
M_t=
\bigl(
\mathcal{D},
\mathcal{P}_t,
\mathcal{B}_t
\bigr),
\]
where $\mathcal{D}$ is a set of $K$ question-conditioned evidence demands,
$\mathcal{P}_t$ records the execution status of the initial acquisition plan,
and $\mathcal{B}_t$ is the evidence board that stores retained observations and interaction outcomes with demand-level provenance.
The plan's goals and demand assignments remain fixed within an episode; its active subtask and completion statuses change as execution proceeds.
EviSeek updates $M_t$ through the three stages detailed in turn below.

All components are implemented through prompting: throughout this paper, functions $F_{\mathrm{dem}}$, $F_{\mathrm{plan}}$, and the policies $\pi_{\mathrm{seek}}$, $\pi_{\mathrm{keep}}$, and $\pi_{\mathrm{ans}}$ denote \textbf{inference-time prediction procedures instantiated by a pretrained vision-language model}, involving no task-specific training or parameter updates.
EviSeek operates through the benchmark observation and action interfaces. In ESI-Bench, it uses RGB, task-exposed action feedback, and relative poses accumulated from camera actions (Appendix~\ref{app:implementation}).

\subsection{Evidence-Demand Planning}

An embodied question specifies the requested information but does not directly reveal the observations required to establish it. 
Given a question $Q$ and the initial observation $o_0$, EviSeek first constructs a set of evidence demands:
\[
\mathcal{D}
=
F_{\mathrm{dem}}(Q,o_0)
=
\{d_1,\ldots,d_K\}.
\]

Each demand $d_i$ specifies an observable acquisition target intended to support answering the question; each corresponds to one information dependency of $Q$ and is constructed to satisfy three properties.
First, it is \emph{question-conditioned}: the demand targets information relevant to resolving $Q$. 
Second, it is \emph{observable}: the demand can be supported by visual observations or by outcomes produced through legal interactions. 
Third, it is \emph{answer-agnostic}: it specifies what should be observed without assuming that a candidate answer is correct. A physical-test goal can name a candidate configuration, such as a stacking order, and request an observation of its outcome.

The demands define observational targets, while the interaction policy determines how to acquire them.
Based on them, EviSeek constructs an initial acquisition plan:
\[
\begin{aligned}
\mathcal{P}_0
&=
F_{\mathrm{plan}}
(Q,\mathcal{D},o_0)
=
\langle u_1,\ldots,u_J\rangle ,
\\
u_j&=(g_j,\rho(j)),
\end{aligned}
\]
where $g_j$ is the subtask's observable acquisition goal and $\rho(j)\in\{1,\ldots,K\}$ indexes the demand $d_{\rho(j)}$ that $u_j$ serves.

A subtask describes what evidence should be obtained rather than how actions should be executed. 
Its realization is delegated to the interaction module, allowing the agent to adapt acquisition according to newly observed scene information.

\subsection{Evidence-Seeking Interaction}

Given an active acquisition subtask $u_j$ with goal $g_j$, EviSeek selects actions that are expected to produce informative evidence. 
At step $t$, the seeker predicts an acquisition intent:
\[
\xi_t
\sim
\pi_{\mathrm{seek}}
(\cdot
\mid
g_j,o_t,W_t),
\]
where $W_t$ denotes the transient working memory of the current subtask, distinct from the retained evidence board $\mathcal{B}_t$.

The intent $\xi_t$ can specify any legal composition of viewpoint-acquisition and physical-intervention actions (Section~\ref{sec:preliminary}) in $\mathcal{G}_{\mathcal{E}}$, which makes the intent set extensible.
The intent space is an interface-level specification rather than a fixed component of the framework: the concrete intent set is instantiated for each target environment, while the acquisition procedure itself remains unchanged.

After executing $\xi_t$, the environment returns a new observation and interface-level outcome:
\[
(o_{t+1},r_{t+1})
=
\operatorname{Exec}_{\mathcal{E}}(\xi_t).
\]

EviSeek then determines whether the resulting information should be retained:
\[
k_{t+1}
\sim
\pi_{\mathrm{keep}}
(
\cdot
\mid
g_j,o_t,\xi_t,o_{t+1},r_{t+1},W_t
),
\]
where $k_{t+1}\in\{0,1\}$ indicates evidence retention.

Only selected results are added to the evidence board. 
Each retained evidence item is represented as:
\[
b_{t+1}
=
(o_{t+1},r_{t+1},\xi_t,u_j,d_{\rho(j)}),
\]
which preserves the observation, the action outcome, the acquisition intent, the originating subtask, and the associated demand.

The evidence board is updated as:
\[
\mathcal{B}_{t+1}
=
\begin{cases}
\mathcal{B}_{t}\cup\{b_{t+1}\},
&
k_{t+1}=1,\\
\mathcal{B}_{t},
&
k_{t+1}=0.
\end{cases}
\]

Unselected observations may still be used temporarily for control but are excluded from the evidence-board answer context. Task-supplied reference images and task-specific answer interfaces are documented in Appendix~\ref{app:implementation}.

\subsection{Evidence-Oriented Memory}

The evidence board stores selected observations and action outcomes with their originating subtask and demand. These links preserve acquisition provenance and organize the answer context by the information sought during exploration.

\paragraph{Plan progression and stopping.}
The seeker can stop the current acquisition subtask. This closes the subtask as it stands, binds its retained items to the associated demand, and advances execution to the next planned goal. A blocked or timed-out subtask is marked as timed out before execution advances. Neither transition requires a retained item to exist or a coverage predicate to be satisfied. The agent answers after the plan has no remaining subtask or the episode budget forces termination. Action selection remains observation-conditioned throughout execution. 

\paragraph{Answer context.}

At the stopping time $\tau$, the final answer is generated from the evidence-oriented memory:
\[
\hat A
\sim
\pi_{\mathrm{ans}}
(
\cdot
\mid
Q,\mathcal D,\mathcal B_\tau
).
\]

The evidence-board answerer receives the task question and answer contract, the evidence demands without coverage-status labels, and the selected evidence record, including applicable task-supplied references. Acquisition plans and the full interaction history are not appended to this context. Task-specific benchmark callbacks can instead invoke their prescribed answer interface; Appendix~\ref{app:implementation} records this boundary. We evaluate this answer context through both answer correctness and visible evidence support.

\begin{table*}[t]
\centering
\normalsize
\setlength{\tabcolsep}{4pt}
\renewcommand{\arraystretch}{1.08}
\begin{tabular*}{\textwidth}{@{\extracolsep{\fill}}lccccccc@{}}
\toprule
\multirow{2}{*}{\textbf{Category}} & \multirow{2}{*}{\textbf{\#Q}} & \multicolumn{3}{c}{\textbf{\qwenname{}}} & \multicolumn{3}{c}{\textbf{Gemini 3.1 Pro}} \\
\cmidrule(lr){3-5}\cmidrule(lr){6-8}
& & Passive & Single-model & EviSeek & Passive & Single-model & EviSeek \\
\midrule
Perceptual Grounding & 750 & 58.0 & 62.3 & \textbf{78.5} & 75.9 & 80.0 & \textbf{84.9} \\
Physical Structure & 254 & \textbf{55.9} & 52.4 & 50.4 & 62.2 & 69.3 & \textbf{70.5} \\
Physical Dynamics & 150 & 32.7 & \textbf{65.3} & 53.3 & 40.0 & 64.0 & \textbf{68.0} \\
Specular Reflection & 300 & \textbf{49.3} & 44.0 & 44.3 & 47.7 & \textbf{54.0} & 49.7 \\
Spatial Relations & 781 & 27.9 & 28.3 & \textbf{46.2} & 27.8 & 36.9 & \textbf{46.6} \\
Metric Comparison & 319 & 30.4 & \textbf{42.0} & 33.5 & 44.5 & \textbf{52.7} & 44.5 \\
Cognitive Mapping & 260 & 54.6 & 53.5 & \textbf{56.2} & \textbf{61.9} & 57.7 & 59.6 \\
Enumerative Perception & 270 & 1.5 & 3.0 & \textbf{7.4} & 4.1 & \textbf{14.8} & 14.4 \\
Temporal Understanding & 275 & 38.2 & 65.1 & \textbf{67.6} & 38.9 & 63.6 & \textbf{72.7} \\
Action Sequencing & 77 & 67.5 & 72.7 & \textbf{79.2} & 32.5 & 75.3 & \textbf{93.5} \\
\midrule
\textbf{Overall} & \textbf{3436} & 40.5 & 45.6 & \textbf{52.7} & 46.4 & 55.7 & \textbf{59.3} \\
\bottomrule
\end{tabular*}
\caption{ESI-Bench accuracy (\%) on 3,436 questions. \#Q is task size; Overall is question-weighted. Bold marks the best result per backbone.}
\label{tab:grounded_delta}
\end{table*}

\section{Experiments}
\label{sec:experiments}

We evaluate whether explicit evidence demands and selective retention improve task success and the evidence available for answering. ESI-Bench tests accuracy, visual support, and resource use across Qwen and Gemini backbones, with Qwen ablations examining the two components. EXPRESS-Bench tests evidence acquisition through navigation with Gemini 3.1 Flash-Lite.

\subsection{Benchmarks and Comparison Protocol}
\label{sec:setup}

\paragraph{Primary benchmark.}
ESI-Bench~\citep{hong2026esibench} contains 10 task families and 29 subcategories in OmniGibson~\citep{li2023behavior}, covering perceptual grounding, spatial and temporal reasoning, navigation, and physical manipulation. We evaluate $N=3{,}436$ questions after the benchmark exclusions described in Appendix~\ref{app:accounting}. The active systems use the same question IDs, scene initializations, task specifications, answer evaluator, and legal atomic action space. Both have a maximum of 30 environment steps, with every executed primitive action charged to the budget.

\paragraph{Backbones.}
We use \qwenname{}~\citep{bai2025qwen3} and \proname{}~\citep{googledeepmind2026gemini31pro} as separate backbones. Within each model family, the same backbone supplies all EviSeek VLM stages and the corresponding baselines. Appendix~\ref{app:accounting} reports model identifiers and decoding parameters.

\paragraph{Compared systems.}
\textbf{\emph{Single-model}} uses the benchmark's single-VLM perception--action loop with task-specific exploration advice. Its memory retains the full textual history and up to five preceding scene views alongside the current view. It uses self-reported answer confidence on a 0--1 scale, with a stopping threshold of 0.80. Appendix~\ref{app:single_model} details its inputs, memory, and stopping rule. \textbf{\emph{EviSeek}} organizes planning, action selection, retention, and answering around evidence demands. It consumes the same task content and action interface, with its own structured outputs and selective evidence board. Execution advances through the initial plan according to subtask completion and the interaction budget. \textbf{\emph{Passive}} answers from a single scene image without interacting, which provides a noninteractive reference on the same question set.

\paragraph{Metrics.}
We measure answer accuracy as
\[
\mathrm{Acc}=\frac{100}{N}\sum_{i=1}^{N}s_i,
\]
where $N$ is the number of evaluated questions and $s_i\in\{0,1\}$ is the benchmark's correctness verdict. We report task-family accuracy and the question-weighted overall result, followed by visual evidence support and episode-level cost. 

\begin{figure}[t]
\centering
\includegraphics[width=\columnwidth]{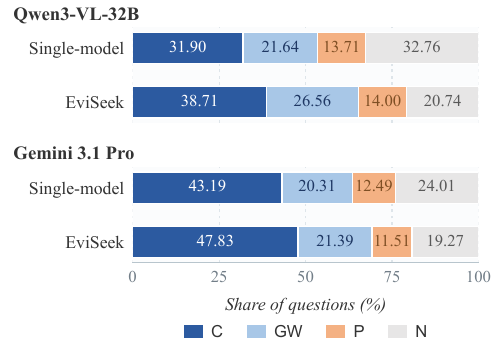}
\caption{Correctness and visual-support decomposition on ESI-Bench (\%). $C$ and $\mathrm{GW}$ are support-weighted correct and incorrect mass; $P$ and $\mathrm{N}$ are their residuals. }
\label{fig:grounded_quadrants}
\end{figure}

\subsection{Task Performance Across Backbones}
\label{sec:q1}

EviSeek improves overall accuracy over Single-model by 7.1 percentage points with the open-weight Qwen backbone and 3.7 points with the closed Gemini backbone (Table~\ref{tab:grounded_delta}). Single-model already interacts using the same backbone, task interface, and step ceiling. The additional gain therefore supports organizing acquisition and retention around explicit evidence needs. Paired tests support EviSeek's gains over both Single-model and Passive for each backbone (Appendix~\ref{app:qwen_pilot}).

The larger gain with Qwen narrows its accuracy gap to Gemini. Gemini retains the highest final accuracy, but its improvement shows that a stronger backbone still benefits from explicit acquisition objectives and selective memory. EviSeek thus complements both tested backbones, with a greater benefit for the weaker baseline.

The shared gains are clearest in tasks that require revealing hidden content or resolving spatial relations, and extend to temporal understanding and action sequencing. Precise measurement remains difficult: Metric Comparison regresses for both backbones, while Physical Dynamics improves only with Gemini.

\subsection{Visual Evidence Supporting the Answers}
\label{sec:q2}

We evaluate answer-time visual support to connect task success with evidence acquisition.

\paragraph{Visual support and correctness decomposition.}
Following EXPRESS-Bench~\citep{jiang2025beyond}, we weight correctness by visual support. GPT-5.5 grades each question's answer-time images with $\delta_i\in\{0,0.5,1\}$: 1 for verifiable decisive evidence, 0.5 for relevant but unresolved evidence, and 0 for absent decisive evidence. The judge receives the question, options, ground-truth answer, and numbered images, with system predictions withheld. Using the notation from Section~\ref{sec:setup}, we define
\[
\begin{aligned}
C &= \frac{100}{N}\sum_i\delta_i s_i,\\
\mathrm{GW} &= \frac{100}{N}\sum_i\delta_i(1-s_i),\\
P &= \mathrm{Acc}-C,\\
\mathrm{N} &= 100-\mathrm{Acc}-\mathrm{GW}.
\end{aligned}
\]
Here $C$ and $\mathrm{GW}$ are support-weighted correct and incorrect mass, with residuals $P$ and $\mathrm{N}$. The components sum to 100\%; partial support splits a question equally between its supported and residual components. Grading includes reference and acquired images but excludes action-result text, so $P$ does not measure reliance on prior knowledge. Appendix~\ref{app:grounded_protocol} details the protocol, and Table~\ref{tab:grounded_c} reports subcategory $C$.

\paragraph{Results.}
Both backbones increase $C$ and reduce $\mathrm{N}$ (Figure~\ref{fig:grounded_quadrants}). This combination links the accuracy gain to more visually supported success and less residual error, supporting evidence acquisition directed toward the question. Qwen's larger reduction in $\mathrm{N}$ accompanies its larger accuracy gain; Gemini benefits from the same pattern at a stronger starting point.

$P$ stays nearly unchanged for Qwen and decreases for Gemini. Both backbones' accuracy gains are therefore concentrated in $C$, tying improved task success to stronger visual support at answer time.

$\mathrm{GW}$ increases even as the total error rate falls, so a larger share of the remaining errors carries visual support. The larger rise in $\mathrm{GW}$ for Qwen shows that its greater improvement in evidence availability is only partly reflected in correct answers. This points to the remaining need to resolve incomplete or ambiguous evidence and interpret it correctly.

Overall visual support, $C+\mathrm{GW}$, increases despite fewer answer-time images (Table~\ref{tab:cost}). EviSeek thus supplies a more useful visual context with fewer images. Together, the four components support the role of evidence demands and the evidence board in directing acquisition toward question-relevant observations and preserving them for answering. The ablations below examine their individual contributions.

\subsection{Interaction Cost and Component Analysis}
\label{sec:analysis}

\paragraph{Episode-level cost.}
EviSeek reduces mean total tokens per episode relative to Single-model by 45.1\% with Qwen and 13.9\% with Gemini, while improving accuracy and visual support (Table~\ref{tab:cost}). Both backbones also use fewer model calls despite EviSeek's multiple VLM stages. Qwen gains more in both accuracy and token efficiency.

Acquisition intents can execute several primitive actions between calls, while selective retention supplies a compact answer context. The savings are consistent with separating environmental action from repeated model reasoning. Gemini illustrates this distinction: it takes more environment steps while using fewer model calls and tokens. The resource advantage therefore concerns model use, without requiring uniformly shorter trajectories.

Calls cover all recorded VLM stages; tokens include input, output, and separately logged reasoning. All episode-level cost statistics are averaged over the same complete set of 3,436 ESI-Bench questions for every system, under the shared 30-step ceiling.

\begin{table}[t]
\centering
\normalsize
\setlength{\tabcolsep}{4pt}
\begin{tabular*}{\columnwidth}{@{\extracolsep{\fill}}lcccc@{}}
\toprule
\textbf{Method} & \textbf{Steps} & \textbf{Calls} & \textbf{Tokens} & \textbf{Images} \\
\midrule
\multicolumn{5}{@{}l}{\textit{\qwenname{}}} \\
Single-model & 18.79 & 20.55 & 139{,}380 & 5.91 \\
EviSeek & 18.11 & 12.69 & 76{,}470 & 4.39 \\
\midrule
\multicolumn{5}{@{}l}{\textit{\proname{}}} \\
Single-model & 10.71 & 12.15 & 86{,}887 & 5.24 \\
EviSeek & 12.17 & 10.05 & 74{,}839 & 4.38 \\
\bottomrule
\end{tabular*}
\caption{Mean resource use across all 3,436 ESI-Bench questions: primitive steps, VLM calls, and total tokens per episode. Images denotes images per answer.}
\label{tab:cost}
\end{table}

\paragraph{Module ablations.}
Table~\ref{tab:components} evaluates three ablations with Qwen on all 3,436 questions. \emph{w/o evidence demands} replaces demand-conditioned planning with direct question-to-subtask decomposition and removes demand descriptions and links from model inputs. We rerun interaction while retaining the action interface, selective evidence board, and stopping rules.

\emph{w/o evidence board} replaces the answer-time board with Single-model memory: all textual history and the current view plus up to five preceding ordinary views. \emph{w/o selective retention} replaces selected board frames with uniformly sampled trajectory frames under the same 20-frame capacity limit. Both memory variants reuse EviSeek's recorded trajectories and rerun only answering, separating the effect of answer-time memory from exploration. Appendix~\ref{app:qwen_pilot} gives replay and scoring details.

\begin{table}[t]
\centering
\normalsize
\begin{tabular*}{\columnwidth}{@{\extracolsep{\fill}}lcc@{}}
\toprule
\textbf{Variant} & \textbf{Acc} & \textbf{$\Delta$Acc} \\
\midrule
EviSeek & 52.7 & --- \\
w/o evidence demands & 46.6 & $-6.1$ \\
w/o evidence board & 47.6 & $-5.1$ \\
w/o selective retention & 47.4 & $-5.3$ \\
\bottomrule
\end{tabular*}
\caption{Qwen ablations on ESI-Bench: accuracy (Acc, \%) and change from the corresponding online or replay reference ($\Delta$Acc, pp).}
\label{tab:components}
\end{table}

Removing evidence demands lowers online accuracy despite retaining task decomposition, supporting explicit informational targets. Both memory variants also reduce accuracy on fixed trajectories. The recent-view result supports preserving earlier evidence, while uniform sampling supports relevance-based selection under the same memory capacity limit. Together, these results support specifying what interaction should establish and selectively preserving the resulting evidence for answering.

\subsection{Navigation Results on EXPRESS-Bench}
\label{sec:q3}

EXPRESS-Bench~\citep{jiang2025beyond} tests whether navigation acquires observations that support free-text answers. It contains $2{,}044$ question--trajectory pairs across 174 HM3D scenes~\citep{ramakrishnan2021habitat}. We map acquisition intents to its legal navigation actions and use RGB observations without ESI-Bench's task-specific guidance. This navigation variant includes coverage-dependent control and replanning, whereas the ESI-Bench configuration follows its initial acquisition plan. Appendix~\ref{app:intents} details the action adaptation.

All interactive methods use Gemini 3.1 Flash-Lite and a maximum of 120 environment steps. Single-model is the RGB baseline. Fine-EQA~\citep{jiang2025beyond} uses frontier-based navigation with RGB-D mapping; we reproduce its official implementation with Flash-Lite for every internal model call. Blind answers from the question alone. The official GPT-4o-mini judge evaluates answers under the benchmark protocol. We report grounded correctness $C$, answer-only correctness $C^\ast$, path-weighted correctness $E_{\mathrm{path}}$, and terminal geodesic distance $d_T$. Appendix~\ref{app:express_details} gives the scoring definitions.

\begin{table}[t]
\centering
\normalsize
\setlength{\tabcolsep}{3pt}
\renewcommand{\arraystretch}{1.05}
\begin{tabular*}{\columnwidth}{@{\extracolsep{\fill}}lcccc@{}}
\toprule
\textbf{Method} & $C\uparrow$ & $C^\ast\uparrow$ & $E_{\mathrm{path}}\uparrow$ & $d_T\downarrow$ \\
\midrule
Blind & --- & 53.79 & --- & --- \\
Single-model & 19.48 & 40.09 & 18.66 & 6.42 \\
Fine-EQA & \textbf{49.92} & \textbf{67.10} & 33.15 & \textbf{5.60} \\
EviSeek (nav.) & 41.60 & 63.09 & \textbf{37.77} & 6.37 \\
\bottomrule
\end{tabular*}
\caption{EXPRESS-Bench with Gemini 3.1 Flash-Lite. }
\label{tab:express_full}
\end{table}

The navigation variant improves both grounded and answer-only correctness over the RGB Single-model agent (Table~\ref{tab:express_full}). The grounded improvement shows that the benefit extends to answers supported by observations acquired during exploration. EviSeek also exceeds the question-only Blind reference in answer quality, whereas Single-model falls below it. This contrast highlights the value of organizing navigation around the information needed for the question: interaction becomes useful when it produces evidence that supports the final answer.

EviSeek approaches Fine-EQA's answer-only correctness and attains the highest path-weighted correctness among the compared systems. This performance is achieved with RGB observations, without depth input or external mapping tools, whereas Fine-EQA uses depth observations and 3D voxel and semantic maps to guide exploration~\citep{jiang2025beyond}. The comparison highlights the value of question-directed evidence acquisition under a more restricted observation interface, while Fine-EQA retains higher grounded correctness and a lower terminal distance.

\section{Conclusion}

EviSeek addresses undirected interaction in active embodied question answering by making the agent's information needs explicit. Its evidence demands connect acquisition goals to selectively retained observations, giving action selection and memory a shared informational purpose. On ESI-Bench, this design improves accuracy and visual support over Single-model with Qwen and Gemini while reducing mean token use; ablations support the contributions of explicit demands and selective retention. The navigation variant also improves grounded answering over the RGB Single-model agent on EXPRESS-Bench without external mapping tools. These findings support organizing embodied interaction around what the agent needs to learn, with actions chosen to reveal relevant evidence and memory preserving it for the final answer.

\section*{Limitations}

EviSeek's evidence board organizes observations without constructing a metric spatial model, leaving spatial inference to the pretrained VLM. Metric Comparison regresses with both backbones, including Gemini 3.1 Pro. Spatial uncertainty may still contribute to undirected interaction despite explicit evidence goals. Explicit spatial modeling and stronger native spatial reasoning are complementary directions to explore.

Recent evaluations report substantial manipulation and navigation capabilities for GPT-6 Astra in their tested settings, alongside limits in precision and dynamic control~\citep{zhang2026unexpected,team2026systematically}. Its performance with EviSeek remains untested. A matched comparison with and without EviSeek would assess the framework's added value.

Evaluation is limited to two simulated benchmarks with predefined action interfaces. Visual domain shifts, sensor noise, and uncertain contact and actuation on physical robots may alter the evidence obtained from planned actions. Transfer of EviSeek's acquisition and retention decisions across this gap remains untested.

\bibliography{references}
\clearpage
\appendix

\section{Evaluation Accounting and Model Configuration}
\label{app:accounting}

\paragraph{Question set.}
ESI-Bench releases $3{,}502$ questions. The evaluation excludes 66 questions: 60 are skipped by the official active-perception code and 6 Unobserved Change questions have reference images that fail to render. The main comparison uses $3{,}436$ question IDs. Each task-family value uses its actual question count, and overall accuracy is a micro-average over all questions. Abstentions, unparseable predictions, and episodes that terminate without an answer count as incorrect.

\paragraph{Answer-format normalization.}
All three Qwen systems use two deterministic normalizations: pipe-separated stacking orders are accepted as comma-separated orders, and complete navigation routes are reduced to blank positions only when all prompt-supplied steps match and each blank maps to a unique listed option. Neither transformation uses ground-truth answers.

\paragraph{Passive input protocol.}
Passive receives one scene image and the question, without interaction history. Qwen Passive uses the pre-action view for each of the 3,436 questions. Its predictions use the same answer-format normalization as the Qwen active systems.

\paragraph{Paired significance tests.}
We compare EviSeek's overall accuracy with Single-model and Passive separately for each backbone, pairing all 3,436 question IDs and counting missing answers as incorrect. The tests use the scores in Table~\ref{tab:grounded_delta}. We apply two-sided exact McNemar tests and Holm correction across the four comparisons. Pointwise 95\% confidence intervals use 200,000 paired bootstrap resamples of question IDs and percentile endpoints (seed 20261012). Table~\ref{tab:main_significance} reports the results.

\paragraph{Qwen3-VL-32B.}
The Qwen backbone is \texttt{Qwen3-VL-32B-Instruct}, served in BF16 through an OpenAI-compatible vLLM endpoint under the alias \texttt{qwen3-vl-32b}. EviSeek and Single-model use temperature 0.2, top-$p$ 0.9, and a ceiling of 30 primitive actions. Passive uses temperature 0, top-$p$ 1, and a 512-token output limit.

\paragraph{Gemini 3.1 Pro.}
The Gemini ESI-Bench experiments use \texttt{gemini-3.1-pro-preview} for EviSeek, Single-model, and Passive. EviSeek requests temperature 0.2, top-$p$ 0.9, and a thinking budget of 512 through a Gemini-compatible endpoint. Both active systems have a ceiling of 30 primitive actions.

\paragraph{Gemini 3.1 Flash-Lite.}
We use this model under the identifier \texttt{gemini-3.1-flash-lite} for all EXPRESS-Bench agents, with a maximum of 120 environment steps. Single-model and EviSeek receive RGB observations; Fine-EQA retains RGB-D mapping and uses Flash-Lite for every internal model call. Flash-Lite also supplies the ESI-Bench intent ablation in Appendix~\ref{app:supplementary}.

\paragraph{GPT judges.}
ESI-Bench visual support is assessed by \texttt{gpt-5.5} through a Responses-compatible API with low reasoning effort and JSON output. The output limit is 4,096 tokens, increased to 8,192 when retrying incomplete outputs; temperature and top-$p$ are not overridden. EXPRESS-Bench uses its official \texttt{gpt-4o-mini} judge for answer and grounding scores.

\paragraph{EXPRESS-Bench scoring.}
\label{app:express_details}

The official GPT-4o-mini judge assigns free-text answers a grounding grade $\delta\in\{0,0.5,1\}$ and an answer score $\sigma\in[1,5]$. Grounded correctness is $C=\overline{\delta\sigma}/5$, and answer-only correctness is $C^\ast=\overline{\sigma}/5$. We also use the benchmark's path-weighted correctness $E_{\mathrm{path}}$ and terminal geodesic distance $d_T$. Table~\ref{tab:express_full} reports the three correctness scores on a 0--100 scale and distance in meters. EXPRESS-Bench grades free-text answers, whereas ESI-Bench uses binary correctness. Their support-weighted scores therefore use different answer-scoring rules and are not directly comparable.

\section{Visual-Support Protocol}
\label{app:grounded_protocol}

\paragraph{Context extraction.}
For each scored prediction, $V_i$ is the image set attached to the model call that produced that prediction. Images available earlier in the trajectory but absent from that call are excluded. Task-supplied reference images are included if attached, with their provenance recorded separately from environment observations. Extraction preserves image order and deduplicates repeated identical paths. For judging, images are converted to RGB, fitted within $1365\times768$ pixels without upscaling, and numbered in an upper-left overlay. JPEG conversion uses quality 90; compressed inputs are saved as PNG after numbering to avoid further lossy compression. The extracted image set also determines the images-per-answer statistic in Table~\ref{tab:cost}.

\paragraph{Judge request and rubric.}
The GPT-5.5 judge uses the configuration in Appendix~\ref{app:accounting}. The request contains the question, options, reference answer, and numbered images, but not the system prediction. Grade 1 requires all decisive facts to be visually verifiable; 0.5 denotes relevant visible content with a decisive fact unresolved; and 0 denotes absent evidence. Task-supplied references and acquired views are equally valid, multiple frames can jointly establish support, and ambiguity between adjacent grades is resolved toward the lower grade. Examples distinguish an ambiguity resolved by a new angle, visible objects whose relative size remains indeterminate, and a physical outcome never observed.

\section{Full Task Breakdown}
\label{app:subcategory}

Tables~\ref{tab:grounded_delta_sub} and~\ref{tab:grounded_c} report answer accuracy and visual-support-weighted accuracy for all 29 subcategories, with the same question IDs and denominators.
\begin{table*}[!htbp]
\centering\footnotesize
\setlength{\tabcolsep}{4pt}
\renewcommand{\arraystretch}{1.06}
\begin{tabular}{@{}lc|ccc|ccc@{}}
\toprule
\multirow{2}{*}{\textbf{Subcategory}} & \multirow{2}{*}{\textbf{\#Q}} & \multicolumn{3}{c|}{\textbf{Qwen}} & \multicolumn{3}{c}{\textbf{Gemini}} \\
\cmidrule(lr){3-5}\cmidrule(lr){6-8}
& & Passive & Single-model & EviSeek & Passive & Single-model & EviSeek \\
\midrule
\multicolumn{8}{@{}l}{\textit{Perceptual Grounding}} \\
View Hallucination & 437 & 55.8 & 59.5 & \textbf{78.9} & 77.6 & 82.8 & \textbf{88.1} \\
Partial Occlusion & 95 & 22.1 & 50.5 & \textbf{86.3} & 54.7 & 74.7 & \textbf{85.3} \\
Material Transparency & 218 & \textbf{78.0} & 72.9 & 74.3 & \textbf{81.7} & 76.6 & 78.4 \\
\addlinespace[2pt]
\multicolumn{8}{@{}l}{\textit{Physical Structure}} \\
Rigid Containment & 20 & \textbf{55.0} & 50.0 & 45.0 & 70.0 & \textbf{80.0} & 60.0 \\
Liquid Volume & 136 & \textbf{69.1} & 61.0 & 57.4 & \textbf{75.7} & \textbf{75.7} & 73.5 \\
Deformable & 98 & 37.8 & 40.8 & \textbf{41.8} & 41.8 & 58.2 & \textbf{68.4} \\
\addlinespace[2pt]
\multicolumn{8}{@{}l}{\textit{Physical Dynamics}} \\
Inclined Plane & 61 & 52.5 & \textbf{80.3} & 73.8 & 59.0 & 75.4 & \textbf{85.2} \\
Stacking \& Stability & 89 & 19.1 & \textbf{55.1} & 39.3 & 27.0 & \textbf{56.2} & \textbf{56.2} \\
\addlinespace[2pt]
\multicolumn{8}{@{}l}{\textit{Specular Reflection}} \\
Reflection Authoring & 99 & \textbf{50.5} & \textbf{50.5} & \textbf{50.5} & 53.5 & \textbf{71.7} & 65.7 \\
Spatial Relations & 113 & \textbf{50.4} & 43.4 & 46.9 & 39.8 & \textbf{46.0} & 35.4 \\
Correspondence & 88 & \textbf{46.6} & 37.5 & 34.1 & \textbf{51.1} & 44.3 & 50.0 \\
\addlinespace[2pt]
\multicolumn{8}{@{}l}{\textit{Spatial Relations}} \\
Linear Alignment & 94 & 53.2 & \textbf{56.4} & 38.3 & 37.2 & 68.1 & \textbf{79.8} \\
Geometric Configuration & 568 & 14.8 & 17.6 & \textbf{43.7} & 16.2 & 23.6 & \textbf{36.1} \\
Physical Contact & 119 & \textbf{70.6} & 57.1 & 64.7 & \textbf{75.6} & \textbf{75.6} & 70.6 \\
\addlinespace[2pt]
\multicolumn{8}{@{}l}{\textit{Metric Comparison}} \\
Dimensional Size & 167 & 10.8 & \textbf{26.9} & 6.6 & 23.4 & \textbf{39.5} & 22.2 \\
Spatial Distance & 152 & 52.0 & 58.6 & \textbf{63.2} & 67.8 & 67.1 & \textbf{69.1} \\
\addlinespace[2pt]
\multicolumn{8}{@{}l}{\textit{Cognitive Mapping}} \\
Connectivity & 60 & 58.3 & 55.0 & \textbf{66.7} & 63.3 & 50.0 & \textbf{65.0} \\
Traversable Passage & 60 & 70.0 & \textbf{75.0} & 60.0 & \textbf{71.7} & 66.7 & 63.3 \\
Regional Boundary & 80 & 50.0 & \textbf{52.5} & \textbf{52.5} & \textbf{63.8} & 58.8 & 56.2 \\
Long-Term Navigation & 60 & 41.7 & 31.7 & \textbf{46.7} & 48.3 & \textbf{55.0} & \textbf{55.0} \\
\addlinespace[2pt]
\multicolumn{8}{@{}l}{\textit{Enumerative Perception}} \\
Counting w Occlusion & 30 & 6.7 & \textbf{16.7} & 13.3 & 13.3 & \textbf{23.3} & 10.0 \\
Spatial Segmentation & 30 & 3.3 & 6.7 & \textbf{10.0} & 13.3 & \textbf{40.0} & 26.7 \\
Merged Observation & 60 & 0.0 & 0.0 & \textbf{1.7} & 0.0 & \textbf{10.0} & 3.3 \\
Category Ambiguity & 60 & 0.0 & 0.0 & \textbf{13.3} & 3.3 & 13.3 & \textbf{16.7} \\
Structural Enclosure & 40 & \textbf{2.5} & \textbf{2.5} & \textbf{2.5} & 2.5 & 7.5 & \textbf{10.0} \\
Illumination Variability & 50 & 0.0 & 0.0 & \textbf{6.0} & 0.0 & 8.0 & \textbf{24.0} \\
\addlinespace[2pt]
\multicolumn{8}{@{}l}{\textit{Temporal Understanding}} \\
Unobserved Change & 142 & 39.4 & 85.9 & \textbf{93.0} & 40.1 & 81.0 & \textbf{96.5} \\
Agent Observation & 133 & 36.8 & \textbf{42.9} & 40.6 & 37.6 & 45.1 & \textbf{47.4} \\
\addlinespace[2pt]
\multicolumn{8}{@{}l}{\textit{Action Sequencing}} \\
Action Order Inference & 77 & 67.5 & 72.7 & \textbf{79.2} & 32.5 & 75.3 & \textbf{93.5} \\
\midrule
\textbf{Overall} & \textbf{3436} & 40.5 & 45.6 & \textbf{52.7} & 46.4 & 55.7 & \textbf{59.3} \\
\bottomrule
\end{tabular}
\caption{Accuracy (\%) across the 29 ESI-Bench subcategories. \#Q is subcategory size; Overall is question-weighted. Bold marks the best result per backbone.}
\label{tab:grounded_delta_sub}
\end{table*}

\begin{table*}[!htbp]
\centering\footnotesize
\setlength{\tabcolsep}{7pt}
\renewcommand{\arraystretch}{1.06}
\begin{tabular}{@{}lc|cc|cc@{}}
\toprule
\multirow{2}{*}{\textbf{Subcategory}} & \multirow{2}{*}{\textbf{\#Q}} & \multicolumn{2}{c|}{\textbf{\qwenname{}}} & \multicolumn{2}{c}{\textbf{\proname{}}} \\
\cmidrule(lr){3-4}\cmidrule(lr){5-6}
& & Single-model & EviSeek & Single-model & EviSeek \\
\midrule
\multicolumn{6}{@{}l}{\textit{Perceptual Grounding}} \\
View Hallucination & 437 & 52.86 & 67.16 & 75.51 & 79.18 \\
Partial Occlusion & 95 & 37.89 & 35.26 & 64.21 & 60.53 \\
Material Transparency & 218 & 56.19 & 69.72 & 65.60 & 69.72 \\
\addlinespace[2pt]
\multicolumn{6}{@{}l}{\textit{Physical Structure}} \\
Rigid Containment & 20 & 30.00 & 32.50 & 47.50 & 45.00 \\
Liquid Volume & 136 & 59.93 & 55.88 & 74.26 & 72.06 \\
Deformable & 98 & 33.16 & 32.14 & 44.39 & 54.08 \\
\addlinespace[2pt]
\multicolumn{6}{@{}l}{\textit{Physical Dynamics}} \\
Inclined Plane & 61 & 78.69 & 60.66 & 74.59 & 84.43 \\
Stacking \& Stability & 89 & 30.90 & 23.60 & 30.34 & 33.15 \\
\addlinespace[2pt]
\multicolumn{6}{@{}l}{\textit{Specular Reflection}} \\
Reflection Authoring & 99 & 43.43 & 40.40 & 61.62 & 46.97 \\
Spatial Relations & 113 & 26.55 & 32.30 & 34.51 & 25.66 \\
Correspondence & 88 & 17.61 & 23.30 & 24.43 & 38.07 \\
\addlinespace[2pt]
\multicolumn{6}{@{}l}{\textit{Spatial Relations}} \\
Linear Alignment & 94 & 35.11 & 36.17 & 64.89 & 78.19 \\
Geometric Configuration & 568 & 6.16 & 18.22 & 15.58 & 23.06 \\
Physical Contact & 119 & 34.45 & 44.96 & 40.76 & 45.80 \\
\addlinespace[2pt]
\multicolumn{6}{@{}l}{\textit{Metric Comparison}} \\
Dimensional Size & 167 & 15.57 & 4.19 & 25.75 & 16.47 \\
Spatial Distance & 152 & 30.26 & 45.39 & 46.71 & 54.28 \\
\addlinespace[2pt]
\multicolumn{6}{@{}l}{\textit{Cognitive Mapping}} \\
Connectivity & 60 & 18.33 & 30.00 & 22.50 & 37.50 \\
Traversable Passage & 60 & 17.50 & 23.33 & 15.00 & 19.17 \\
Regional Boundary & 80 & 33.12 & 35.62 & 40.62 & 44.38 \\
Long-Term Navigation & 60 & 10.83 & 23.33 & 20.00 & 25.83 \\
\addlinespace[2pt]
\multicolumn{6}{@{}l}{\textit{Enumerative Perception}} \\
Counting w Occlusion & 30 & 16.67 & 11.67 & 13.33 & 8.33 \\
Spatial Segmentation & 30 & 5.00 & 10.00 & 26.67 & 20.00 \\
Merged Observation & 60 & 0.00 & 0.00 & 7.50 & 2.50 \\
Category Ambiguity & 60 & 0.00 & 10.00 & 12.50 & 15.00 \\
Structural Enclosure & 40 & 1.25 & 2.50 & 3.75 & 7.50 \\
Illumination Variability & 50 & 0.00 & 6.00 & 7.00 & 23.00 \\
\addlinespace[2pt]
\multicolumn{6}{@{}l}{\textit{Temporal Understanding}} \\
Unobserved Change & 142 & 84.15 & 91.55 & 76.06 & 94.01 \\
Agent Observation & 133 & 36.47 & 27.82 & 28.57 & 39.85 \\
\addlinespace[2pt]
\multicolumn{6}{@{}l}{\textit{Action Sequencing}} \\
Action Order Inference & 77 & 15.58 & 74.03 & 62.34 & 83.12 \\
\midrule
\textbf{Overall} & \textbf{3436} & 31.90 & 38.71 & 43.19 & 47.83 \\
\bottomrule
\end{tabular}
\caption{Visual-support-weighted accuracy $C$ (\%) across ESI-Bench subcategories. Correct answers are weighted by answer-time visual support (0, 0.5, or 1). \#Q is subcategory size; Overall is question-weighted.}
\label{tab:grounded_c}
\end{table*}
\section{Single-model Baseline Details}
\label{app:single_model}
Single-model denotes ESI-Bench's single-VLM active agent, evaluated with the Qwen and Gemini backbones. This appendix describes its task inputs, memory, and confidence-based stopping. Model and decoding settings appear in Appendix~\ref{app:accounting}.

\paragraph{Task prompts.}
ESI-Bench supplies the question, candidate answers or answer format, legal actions, and task-specific exploration advice. The baseline thus receives guidance about useful observations and interventions, in addition to the question. Appendix~\ref{app:prompts} reproduces a representative task prompt and the EviSeek stage excerpts.

\paragraph{Memory and per-step inputs.}
The textual history accumulates past actions and outcomes together with preceding model responses. The ordinary visual context contains the current RGB view and available views from the five most recent history steps, giving at most six ordinary scene views. The five-step window limits visual history; it does not truncate the textual history. Task-supplied reference images and additional observations from physical interactions are included when applicable. Single-model does not organize this history into a demand-indexed evidence board.

\paragraph{Task-exposed action feedback.}
Both active systems receive structured execution results in the task history. Fields include \texttt{operation}, \texttt{success}, \texttt{reason}, and task-dependent states such as \texttt{current\_stack}. These can convey physical information beyond RGB: stacking feedback reports placement success after simulator settling and stability checks. The common history serializer does not forward the action result's world-position, quaternion, or \texttt{gt\_orders} fields; final-answer correctness is an evaluation output. EviSeek extracts this shared feedback and associates it with retained observations, as detailed in Appendix~\ref{app:implementation}.

\paragraph{Outputs and confidence.}
The model is prompted to return one JSON object with an action, a tentative answer, a brief rationale, and a scalar \texttt{confidence} in $[0,1]$. Confidence is generated by the VLM as part of its response. It is neither a token probability nor an externally calibrated estimate of answer correctness. The benchmark's threshold is $\theta=0.80$. This online control signal is separate from the post-hoc visual-support grade $\delta$ in Section~\ref{sec:q2}; the support judge does not decide when exploration stops.

\paragraph{Task-dependent stopping.}
Confidence-based stopping requires the reported value to reach $\theta=0.80$ and satisfy the task's stopping checks, which may also involve answer validity and a minimum exploration length. The benchmark retains task-specific stopping logic. An episode can also end through a task-supported \texttt{stop} action or at the 30-step interaction limit.

\paragraph{Final answer.}
At termination, the benchmark resolves an answer from the interaction history. If the response cannot be mapped to the listed options, its forced-choice fallback requests a final selection without further environment interaction.

\paragraph{Task text shared with \projname{}.}
\projname{} receives the same task definition, exploration advice, and action list, distributed across planning, seeking, and answering in each stage's input format. It uses structured stage outputs and subtask-based progression in place of the Single-model action-output template and confidence-based control. Both systems use the task's answer parser and scorer. The comparison changes planning, visual retention, answer context, and stopping under a shared action ceiling.

\section{EviSeek Implementation Details}
\label{app:implementation}

The framework uses prompted VLM operators for demand decomposition, action selection, retention, and answer generation. In ESI-Bench, the acquisition plan is fixed at the start of an episode, while action selection adapts to observations. A seeker stop, blocked subtask, or subtask timeout advances execution to the next acquisition goal. Retained observations remain linked to the originating demand. The answerer is called when the plan ends or the interaction budget is exhausted.

\paragraph{Action-derived relative poses.}
In ESI-Bench, the \texttt{pose} and \texttt{d(east,north,height)} annotations in Appendix~\ref{app:prompts} are internal estimates initialized at zero. Camera commands update nominal translation by 0.25\,m and rotation by $15^\circ$, subject to vertical and pitch bounds. The E/N labels denote internal axes, not world coordinates. No simulator camera pose is read to produce or correct these estimates, so they can differ from actual displacement. For example, a recorded \texttt{d(+0.00E,+0.00N,z0.75)} follows three upward commands. Single-model receives the action history rather than these computed pose annotations.

\paragraph{Feedback annotations.}
EviSeek extracts the latest \texttt{result=} entry from the task history, removes model reasoning, and truncates the feedback to 220 characters. It adds this text to the action trail and attaches a shortened summary to retained outcome frames. A recorded example is \texttt{[ENV] pour\_small\_obj success=True}: the task routine reports that the object cleared the container rim, not that the final answer is correct. Thus, the prompt's ``verdict'' refers to task-exposed execution feedback. Its source is shared with Single-model; its retention and presentation differ.

\paragraph{Task-specific answer interfaces.}
The general answer interface supplies the question and answer contract, evidence demands, selected observations and their notes, and applicable reference images. In Inclined Plane, the benchmark's \texttt{put} action instead invokes an answer call using the resulting physics-frame sequence. Visual-support evaluation uses the images attached to the scored answer call in either interface. Task parsers, forced-choice fallbacks, and scorers follow the benchmark protocol.

Appendix~\ref{app:prompts} presents the stage prompt excerpts.

\section{Acquisition-Intent Instantiation}
\label{app:intents}

\paragraph{Atomic action interfaces.}
Table~\ref{tab:action_space} compares the atomic actions in ESI-Bench (ESI) and EXPRESS-Bench (EXP).
The two interfaces share forward translation, horizontal rotation, and answer submission, while ESI-Bench additionally provides backward, lateral, and vertical translation, pitch control, and per-family manipulation commands; EXPRESS-Bench is navigation-only at a fixed camera height.
Translation steps are $0.25$\,m in both environments; turns are $15^\circ$ in ESI-Bench and $30^\circ$ in EXPRESS-Bench.

\begin{table}[H]
\centering
{\footnotesize
\setlength{\tabcolsep}{4pt}
\renewcommand{\arraystretch}{1.05}
\begin{tabular}{@{}lcc@{}}
\toprule
\textbf{Action} & \textbf{ESI} & \textbf{EXP} \\
\midrule
\multicolumn{3}{@{}l}{Locomotion} \\
\texttt{move\_forward} & \checkmark & \checkmark \\
\texttt{move\_backward} & \checkmark & --- \\
\texttt{move\_left} / \texttt{move\_right} & \checkmark & --- \\
\texttt{move\_up} / \texttt{move\_down} & \checkmark & --- \\
\addlinespace
\multicolumn{3}{@{}l}{Perception} \\
\texttt{turn\_left} / \texttt{turn\_right} & \checkmark & \checkmark \\
\texttt{turn\_up} / \texttt{turn\_down} & \checkmark & --- \\
\addlinespace
\multicolumn{3}{@{}l}{Manipulation} \\
\texttt{pick\_up} obj & \checkmark & --- \\
\texttt{put} obj \texttt{inside} obj & \checkmark & --- \\
\texttt{put} obj \texttt{on} obj & \checkmark & --- \\
\texttt{fill} obj \texttt{with water} & \checkmark & --- \\
\texttt{pour} obj \texttt{from} obj \texttt{to} obj & \checkmark & --- \\
\addlinespace
\multicolumn{3}{@{}l}{Terminal} \\
\texttt{answer}($\hat{y}$, $c$) & \checkmark & \checkmark \\
\bottomrule
\end{tabular}}
\caption{Atomic actions supported by ESI-Bench (ESI) and EXPRESS-Bench (EXP). Checkmarks indicate availability; dashes indicate unsupported actions.}
\label{tab:action_space}
\end{table}

\paragraph{Instantiated intents.}
Table~\ref{tab:intents} lists the viewpoint-acquisition intents instantiated over these two interfaces; physical-intervention intents invoke the manipulation commands of Table~\ref{tab:action_space} directly.
In Table~\ref{tab:intents}, $n$ and $k$ are bounded integers and brackets indicate an optional prefix. Each intent compiles into a short parameterized sequence of atomic actions, clamped to the remaining step budget; every compiled action is executed in the environment and charged one step, so EviSeek operates under exactly the per-primitive step accounting of each benchmark.
During execution, target-tracking intents (APPROACH, ORBIT) may be aborted when their target is lost, whereas geometry-complete intents (e.g., SWEEP) run to completion.

\paragraph{Per-environment adaptation.}
For EXPRESS-Bench, acquisition intents are adapted to its navigation interface.
Intents whose compiled sequences require unavailable primitives (ORBIT, RISE\_TOPDOWN, RETREAT\_WIDE) are not instantiated; SWEEP drops its tilt-down prefix; motion parameters are recalibrated to the $30^\circ$ turn quantum; and GOTO's translation caps are extended so that a single reposition can cross room-scale distances in the larger multi-room scenes. The navigation variant uses RGB observations without depth input or external mapping tools.
The navigation controller uses demand coverage to guide replanning and stopping, whereas the ESI-Bench controller follows its initial acquisition plan.
\begin{table*}[!htbp]
\centering
{\footnotesize
\setlength{\tabcolsep}{6pt}
\renewcommand{\arraystretch}{1.05}
\begin{tabular}{@{}lllcc@{}}
\toprule
\textbf{Intent} & \textbf{Description} & \textbf{Compiled actions} & \textbf{ESI} & \textbf{EXP} \\
\midrule
FACE & Orient toward a bearing & $n \times \texttt{turn\_left/right}$ & \checkmark & \checkmark \\
APPROACH & Close in on a visible target & $k \times \texttt{move\_forward}$ & \checkmark & \checkmark \\
GOTO & Reposition to a seen landmark & $[\texttt{turns}] + k \times \texttt{move\_forward}$ & \checkmark & \checkmark \\
SWEEP & Scan the surroundings in a bounded fan & $[\texttt{turn\_down}] + n \times \texttt{turn\_left/right}$ & \checkmark & \checkmark \\
ORBIT & Shift azimuth around a visible target & $n \times (\texttt{move\_right}, \texttt{turn\_left})$ & \checkmark & --- \\
RISE\_TOPDOWN & Rise for an overhead layout view & $3 \times \texttt{move\_up} + n \times \texttt{turn\_down}$ & \checkmark & --- \\
RETREAT\_WIDE & Widen the field of view & $k \times \texttt{move\_backward}$ & \checkmark & --- \\
\bottomrule
\end{tabular}}
\caption{Viewpoint intents and their compiled atomic actions in ESI-Bench (ESI) and EXPRESS-Bench (EXP). $n$ and $k$ are bounded repetition counts; brackets denote optional prefixes. Checkmarks indicate availability.}
\label{tab:intents}
\end{table*}

\section{Significance Tests and Ablation Protocols}
\label{app:qwen_pilot}

\paragraph{Main comparisons.}
Table~\ref{tab:main_significance} reports EviSeek's accuracy gains over each reference under the paired testing protocol in Appendix~\ref{app:accounting}.

\begin{table*}[!htbp]
\centering\small
\setlength{\tabcolsep}{2pt}
\renewcommand{\arraystretch}{1.1}
\begin{tabular*}{\textwidth}{@{\extracolsep{\fill}}cc|cccc|cccc@{}}
\toprule
\multirow{2}{*}{\textbf{Baseline}} & \multirow{2}{*}{\textbf{Metric}} & \multicolumn{4}{c|}{\textbf{Qwen3-VL-32B}} & \multicolumn{4}{c}{\textbf{Gemini 3.1 Pro}} \\
\cmidrule(lr){3-6}\cmidrule(l){7-10}
& & $\Delta$ & \textbf{95\% CI} & $n_{+}/n_{-}$ & $p$ & $\Delta$ & \textbf{95\% CI} & $n_{+}/n_{-}$ & $p$ \\
\midrule
Passive & Acc & $+12.19$ & $[+10.30,+14.09]$ & $783\,/\,364$ & $<0.001$ & $+12.98$ & $[+11.20,+14.78]$ & $742\,/\,296$ & $<0.001$ \\
Single-model & Acc & $+7.10$ & $[+5.24,+8.99]$ & $673\,/\,429$ & $<0.001$ & $+3.67$ & $[+2.01,+5.36]$ & $496\,/\,370$ & $<0.001$ \\
\bottomrule
\end{tabular*}
\caption{EviSeek versus each baseline on 3,436 paired ESI-Bench questions. $\Delta$ is the accuracy gain (pp), with pointwise paired-bootstrap 95\% CIs. $n_{+}/n_{-}$ counts questions correct only for EviSeek/only for the baseline. Two-sided exact McNemar $p$ values use Holm correction across the four comparisons.}
\label{tab:main_significance}
\end{table*}

\paragraph{Full-benchmark ablations.}
For the ablations in Section~\ref{sec:analysis}, demand removal is compared with the online EviSeek reference. The memory variants are compared with an evidence-board replay using the same recorded trajectories and answer model; they change only the general answer interface described in Appendix~\ref{app:implementation}. The reported $\Delta$Acc is the difference from the corresponding reference. All conditions use the same 3,436 questions and answer-format normalization (Appendix~\ref{app:accounting}), with missing answers counted as incorrect.

\section{Supplementary Experiments}
\label{app:supplementary}

\subsection{Intent Interface Under Flash-Lite}

This experiment uses Gemini 3.1 Flash-Lite on all 3,436 ESI-Bench questions, with demand-coverage checks and replanning. The ablation replaces intent selection with one atomic action per decision while keeping demand planning, retention, answering, and the 30-step budget fixed. Table~\ref{tab:ablation-intent-overall} reports accuracy, tokens per environment step, and mean trajectory length. The accuracy difference measures the contribution of intent selection within this control setting.
\begin{table}[H]
\centering
{\small
\setlength{\tabcolsep}{3pt}
\renewcommand{\arraystretch}{1.1}
\begin{tabular}{@{}lccc@{}}
\toprule
\textbf{Configuration} & \textbf{Acc} & \textbf{Tokens / step} & \textbf{Expl.\ steps} \\
\midrule
Single-model & 46.71 & 6,127 & 9.3 \\
\textsc{EviSeek$-$intent} & 50.84 & 12,863 & 13.3 \\
EviSeek & \textbf{53.08} & 5,591 & 17.5 \\
\bottomrule
\end{tabular}}
\caption{ESI-Bench intent ablation with Gemini 3.1 Flash-Lite. Acc is accuracy (\%); tokens/step uses executed environment steps. Expl.\ steps is mean exploration length.}
\label{tab:ablation-intent-overall}
\end{table}

\section{Qualitative Case Studies}
\label{app:case_studies}

\begin{figure*}[p]
\centering
\includegraphics[width=0.99\textwidth]{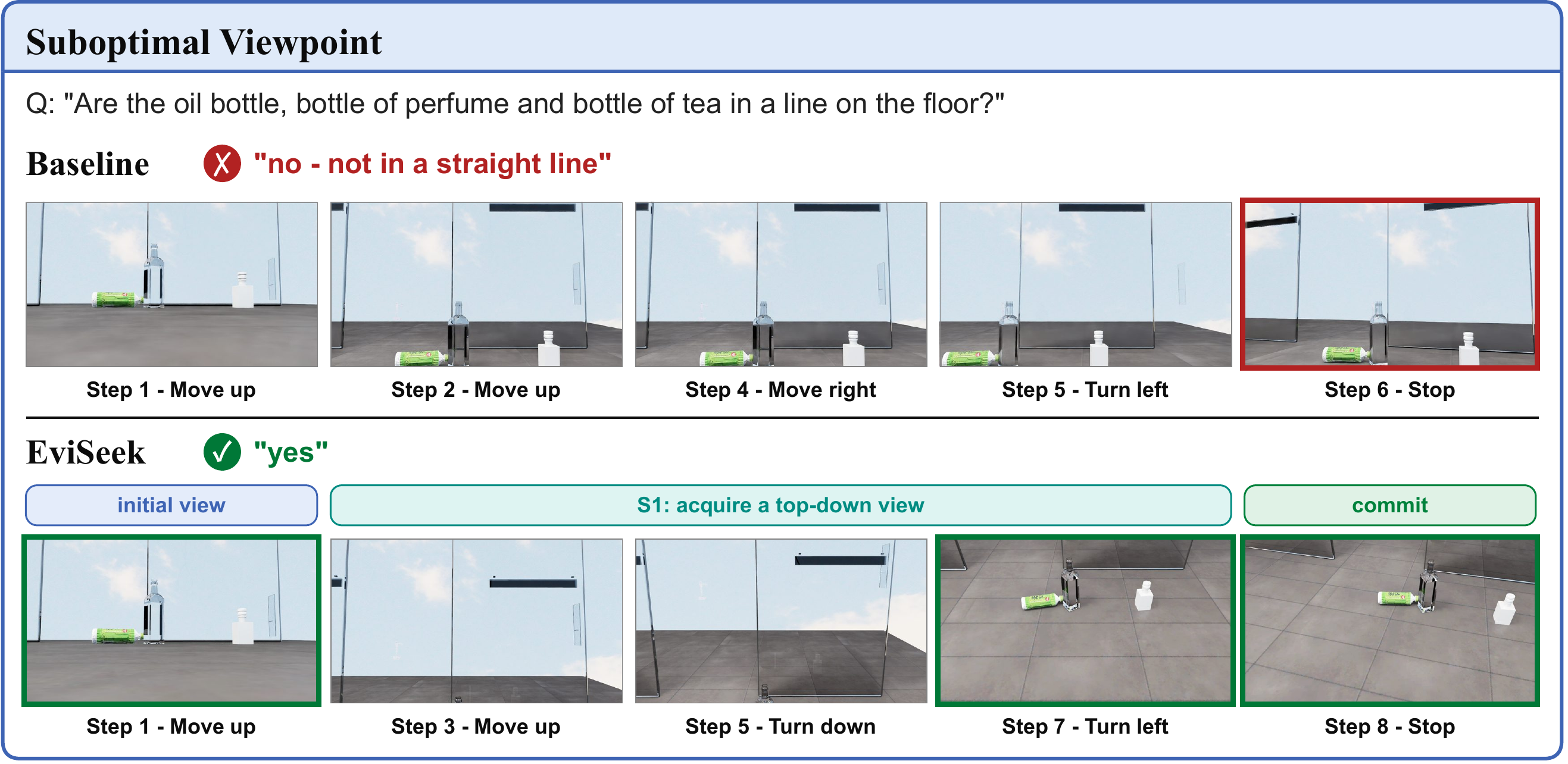}
\caption{Spatial alignment: a top-down acquisition goal reveals that the three objects are collinear. Red frames show Single-model's final views; green frames show retained EviSeek evidence. Colored bars mark acquisition subtasks.}
\label{fig:case_study}
\label{fig:case_study_viewpoint}
\end{figure*}

\begin{figure*}[p]
\centering
\includegraphics[width=0.99\textwidth]{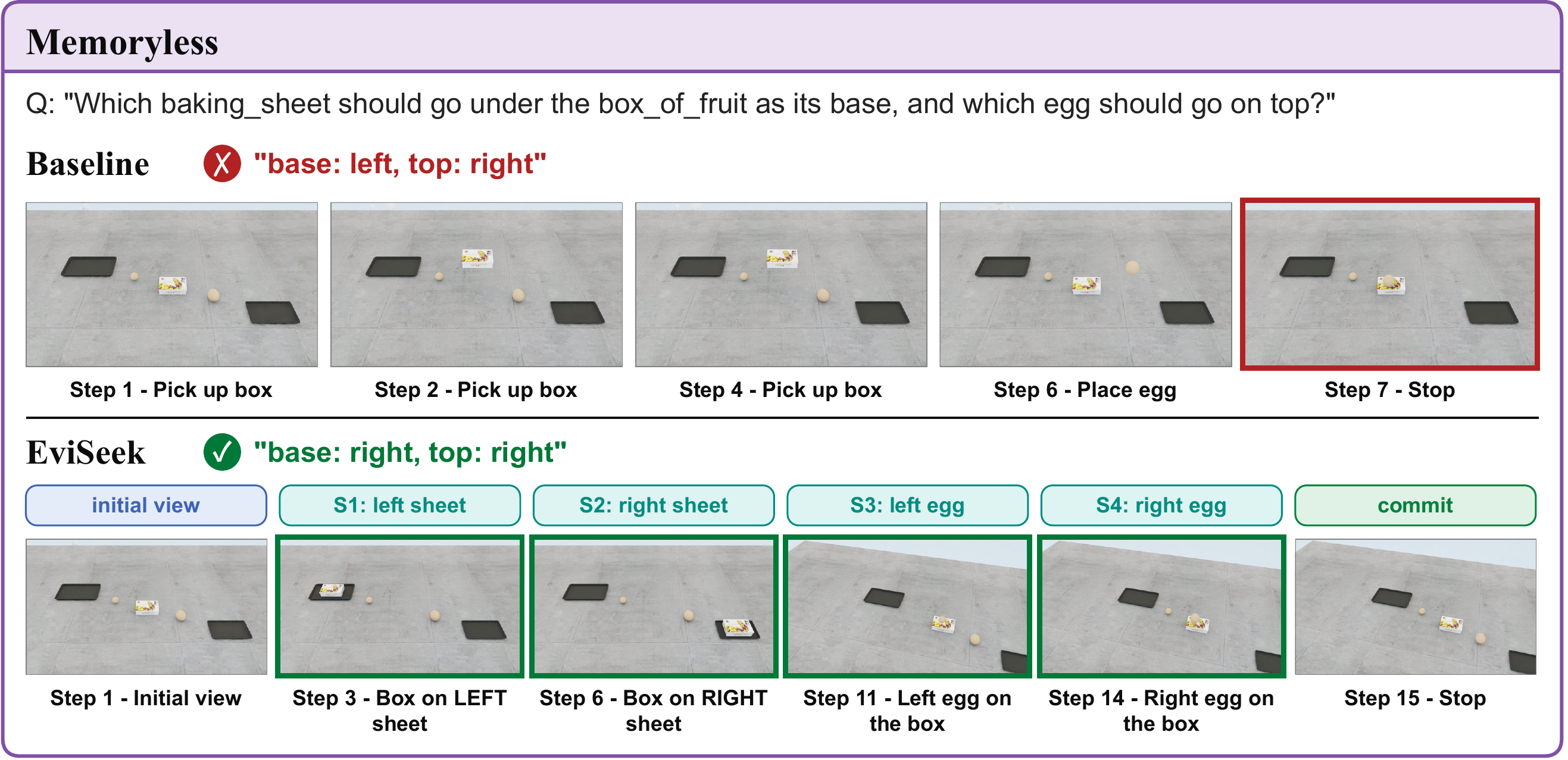}
\caption{Physical stacking: retained observations record the outcomes of testing candidate supports and top objects across manipulations. Red frames show Single-model's final views; green frames show retained EviSeek evidence. Colored bars mark acquisition subtasks.}
\label{fig:case_study_memory}
\end{figure*}

\begin{figure*}[p]
\centering
\includegraphics[width=0.99\textwidth]{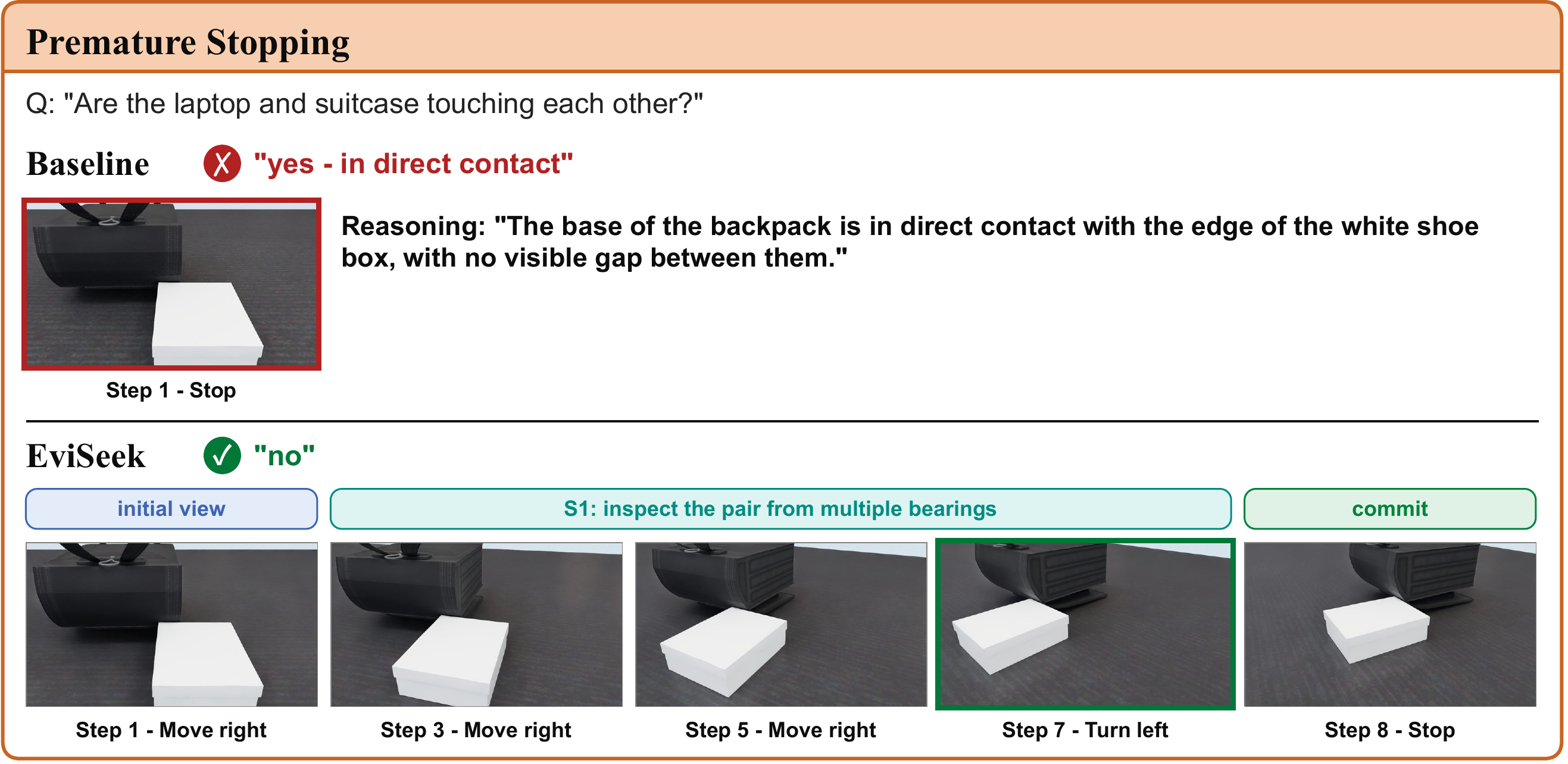}
\caption{Contact judgment: Single-model stops at the initial view, while EviSeek inspects additional bearings before answering. Red frames show Single-model's final views; green frames show retained EviSeek evidence. Colored bars mark acquisition subtasks.}
\label{fig:case_study_stopping}
\end{figure*}

\begin{figure*}[p]
\centering
\includegraphics[width=0.99\textwidth]{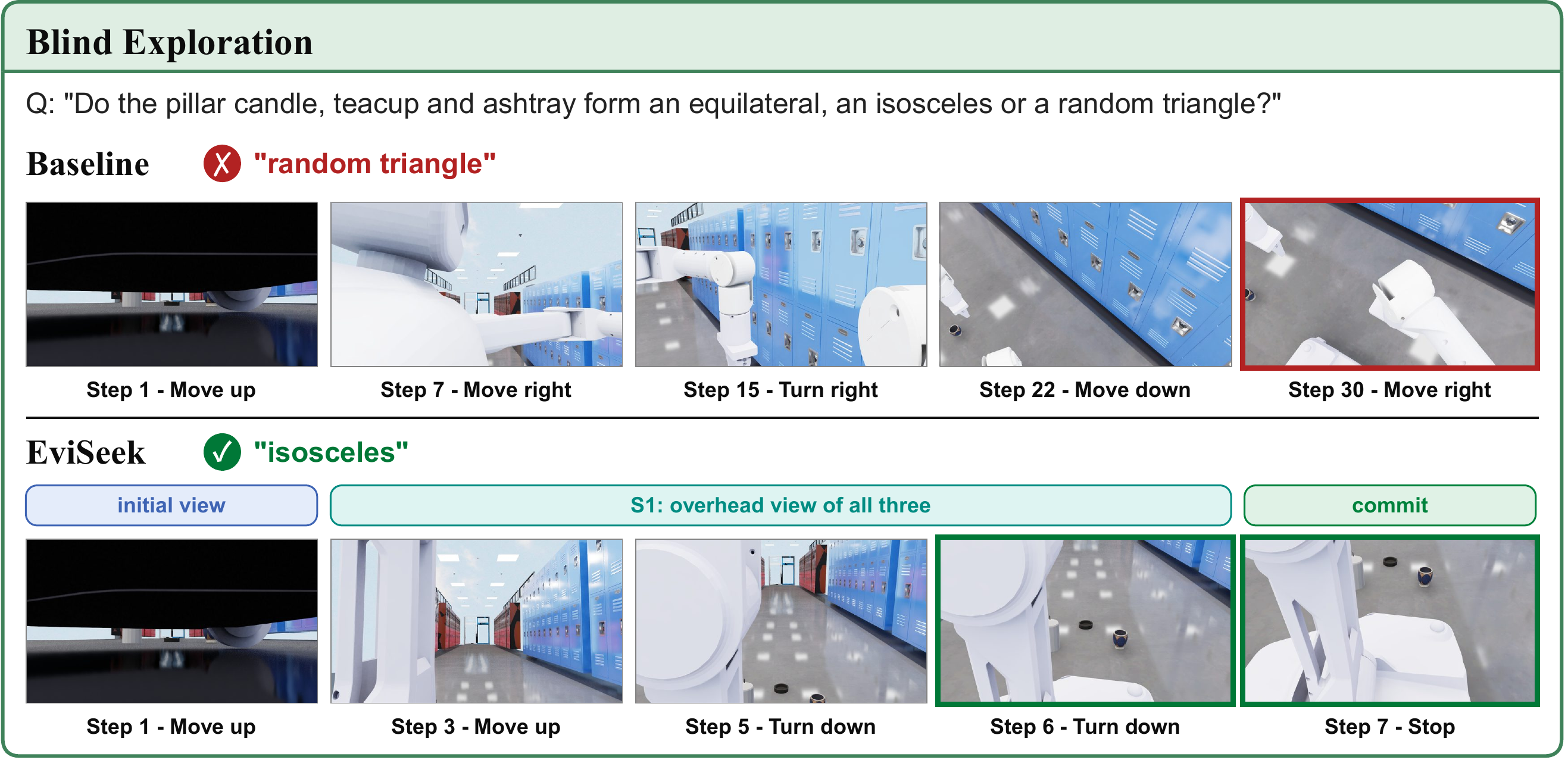}
\caption{Geometric configuration: a targeted overhead view reveals the arrangement of all three objects. Red frames show Single-model's final views; green frames show retained EviSeek evidence. Colored bars mark acquisition subtasks.}
\label{fig:case_study_exploration}
\end{figure*}

Figures~\ref{fig:case_study_viewpoint}--\ref{fig:case_study_exploration} compare Single-model and EviSeek in four ESI-Bench episodes using Gemini 3.1 Flash-Lite; Single-model uses the preview model in the physical-contact case. EviSeek uses demand-coverage checks for subtask completion and stopping. The panels illustrate viewpoint selection, evidence retention, contact verification, and spatial exploration. Step labels indicate positions in each episode.

\clearpage

\clearpage

\clearpage
\section{Prompt Excerpts}
\label{app:prompts}

\paragraph{Task prompts.}
ESI-Bench provides task-specific prompts containing the question, answer format, legal actions, and guidance for obtaining informative observations. We reproduce one representative Material Transparency excerpt below. It illustrates how the official prompt distinguishes containment from side-view overlap and suggests observations or interventions that can resolve the question. For the remaining task prompts and complete answer contracts, we refer readers to ESI-Bench~\citep{hong2026esibench} and its official task implementations.

\begin{promptbox}{Single-model Prompt Excerpt: Material Transparency}
\begin{lstlisting}[numbers=none,frame=none,xleftmargin=0pt,basicstyle=\small\ttfamily,columns=fullflexible,keepspaces=true,breaklines=true,breakatwhitespace=true,breakindent=1em,aboveskip=0pt,belowskip=0pt]
You are a meticulous spatial reasoning expert.
Your goal is to determine if the cherry is inside the round bottom flask.

STRICT CONFIDENCE RULES:
1. PROXIMITY IS NOT CONTAINMENT: Side-view occlusion is a trap. If the object is behind the container, it can look like it is inside.
2. Do not output confidence above 0.5 for yes unless you have strong proof.
3. Proof requires seeing the base of the cherry sitting on the interior bottom surface of the round bottom flask from a top-down angle.
4. DISPROOF SEARCH: Look for a gap. If you see any space between the objects, answer no with high confidence.
[...]
7. ACTION ADVICE: Use move_up and turn_down to look into the container.

You may also use these physical actions:
- pickup: lift the container upward to reveal whether the small object moves with it or remains separate.
- pour: move the small object upward and out past the container rim to test containment.
[...]
Confidence threshold to stop: 0.80.
\end{lstlisting}
\end{promptbox}

\paragraph{Stage prompt excerpts.}
The following boxes show the role prompts for evidence-demand planning, evidence-seeking interaction, evidence retention, and final answering. The excerpts omit task-specific guidance and action schemas; \texttt{[...]} marks omitted text. Appendix~\ref{app:implementation} explains subtask progression and the pose and feedback annotations.

Planning derives evidence demands and an acquisition plan from the task and initial observation. The seeker selects actions and decides whether to retain the current view; the curator selects intermediate frames after an action sequence. The final-answer prompt uses the question, evidence demands, and retained evidence. Task-specific answer interfaces are described in Appendix~\ref{app:implementation}.

\begin{promptbox}{EviSeek Prompt: Evidence-Demand Planning}
\begin{lstlisting}[numbers=none,frame=none,xleftmargin=0pt,basicstyle=\small\ttfamily,columns=fullflexible,keepspaces=true,breaklines=true,breakatwhitespace=true,breakindent=1em,aboveskip=0pt,belowskip=0pt]
## MODE = DECOMPOSE  (called once, at the start)
Do NOT answer the question. Decide what OBSERVATIONS would answer it.
1. EVIDENCE NEEDS: list the distinct pieces of observation the answer requires, as E1, E2, ... Each is a thing you must SEE to be sure - never something you can only infer.
[...]
2. PLAN: 2-5 subtasks, each one ACQUIRING one need. `goal` is an observation to OBTAIN - a specific view to reach, or a manipulation outcome to produce - NOT an action list.
[...]
A comparison or a both-directions test = ONE subtask PER side/direction, each with a single showable outcome.
[...]
Order subtasks so the cheapest discriminating observation is first. If the opening view is flagged degraded, make S1 "reach any informative vantage".
[...]
3. FIND FIRST: when the question names specific object(s) and the opening view does NOT show them, the first subtask must be to FIND the target object(s) - get them into view before anything else.
\end{lstlisting}
\end{promptbox}
\begin{promptbox}{EviSeek Prompt: Evidence-Seeking Interaction}
\begin{lstlisting}[numbers=none,frame=none,xleftmargin=0pt,basicstyle=\small\ttfamily,columns=fullflexible,keepspaces=true,breaklines=true,breakatwhitespace=true,breakindent=1em,aboveskip=0pt,belowskip=0pt]
You do NOT answer the question - it is context so you recognize the goal observation when you see it.
[...]
1. LIGHTNING STOP. If the CURRENT frame already shows the subtask's goal observation, keep it ("keep": true) and STOP NOW - do not buy another view for a goal you are looking at.
[...]
- `keep` (REQUIRED with a maneuver or manip): is the CURRENT frame itself worth keeping as evidence - does it actually show a queried object, a required state, or a view the question needs? true pins THIS frame onto the evidence board the final answer reads; false leaves it. Judge only what is visible in this frame, never what the next maneuver might reveal.
[...]
- Trust the [ENV: ...] lines in the history over what the render seems to show.
\end{lstlisting}
\end{promptbox}

\begin{promptbox}{EviSeek Prompt: Evidence Retention}
\begin{lstlisting}[numbers=none,frame=none,xleftmargin=0pt,basicstyle=\small\ttfamily,columns=fullflexible,keepspaces=true,breaklines=true,breakatwhitespace=true,breakindent=1em,aboveskip=0pt,belowskip=0pt]
# Role: Evidence Curator (one bounded look at a finished maneuver's pass-through frames)
A maneuver just finished. While it executed, the camera passed through the frames below - the Director never saw most of them at decision time. Your ONLY job: pick which of these frames deserve a place on the evidence board.

Input: the subtask's goal observation, the maneuver kind, and the pass-through frames (each headed [MANEUVER FRAME fN @sM | pose]).

Output EXACTLY one JSON object and nothing else:
{"reasoning": "<one or two sentences: what the frames show, which qualify and why>",
"keep_fids": [<frame ids worth keeping, e.g. 12, 15>]}

Rules:
- Judge each frame by what IT shows: a queried object, a required state, or a countable instance. The subtask's goal defines what counts.
- PRECISION over recall: if you cannot name what a frame contributes, do not list it. When in doubt, leave it out - a wrong keep can evict real evidence.
- Skip near-duplicates: several frames of the SAME view from almost the same pose contribute one piece of evidence - list the clearest single frame.
- An empty list ("keep_fids": []) is a normal answer when nothing qualifies.
- No narration beyond `reasoning`. No plans. No actions.
\end{lstlisting}
\end{promptbox}
\newpage
\begin{promptbox}{EviSeek Prompt: Final Answering}
\begin{lstlisting}[numbers=none,frame=none,xleftmargin=0pt,basicstyle=\small\ttfamily,columns=fullflexible,keepspaces=true,breaklines=true,breakatwhitespace=true,breakindent=1em,aboveskip=0pt,belowskip=0pt]
# Role: Answerer
You give the final answer to a question about a 3D scene, on behalf of an embodied agent that has finished exploring it. Your input is the question, the answer rules, and the EVIDENCE BOARD - the frames the agent kept as evidence, each headed by the action that produced it (plus the simulator's [ENV] verdict when that action had one), the camera position d(east,north,height) it was taken from, and a one-line caption - plus the list of evidence needs. You see no planning text and nobody else's reasoning. Answer strictly from what the board's frames actually show; the captions are labels, the pixels are the evidence. Return EXACTLY one JSON object and nothing else.
- answer: repeat one option EXACTLY as listed in the question block. Multi-part questions: one choice per part, in the order asked, joined with " | ". Never "not sure".
- confidence (0.0-1.0): how strongly the board supports the answer. Reported only.

Output: {"answer": "<option exactly as listed>", "confidence": 0.0, "reasoning":
"<brief>"}
\end{lstlisting}
\end{promptbox}

\end{document}